\UseRawInputEncoding
\documentclass{article}

\usepackage[accepted]{icml2024}

\usepackage{microtype}
\usepackage{graphicx}
\usepackage{booktabs}
\usepackage{subcaption}
\usepackage{caption}
\usepackage[table]{xcolor}
\usepackage{array}
\usepackage{multirow}
\usepackage{float}
\usepackage{placeins}
\usepackage{fancyvrb}
\usepackage{listings}
\usepackage{color,soul}
\usepackage{tikz}
\usetikzlibrary{positioning, fit, backgrounds, arrows.meta}
\usetikzlibrary{positioning, arrows.meta}
\usepackage[colorlinks=true,linkcolor=blue,urlcolor=blue,citecolor=blue,anchorcolor=blue]{hyperref}
\usepackage{url}

\usepackage{float}
\usepackage[most]{tcolorbox}
\tcbuselibrary{listings,breakable}

\newtcblisting{promptbox}{
  breakable,
  listing only,
  colback=black!3,
  colframe=black!30,
  boxrule=0.4pt,
  arc=1pt,
  left=4pt,
  right=4pt,
  top=4pt,
  bottom=4pt,
  width=\columnwidth,
  listing options={
    basicstyle=\scriptsize\ttfamily,
    breaklines=true,
    breakatwhitespace=false,
    columns=fullflexible,
    keepspaces=true,
    showstringspaces=false,
    tabsize=2
  },
  before skip=4pt,
  after skip=4pt
}
\usepackage[dvipsnames]{xcolor}
\usepackage{tabularx}

\usepackage{amsmath}
\usepackage{amssymb}
\usepackage{mathtools}
\usepackage{amsthm}
\usepackage{enumitem}
\theoremstyle{plain}

\theoremstyle{definition}

\theoremstyle{remark}

\icmltitlerunning{\textsc{DeepSciVerify}}

\begin{document}

\twocolumn[
\icmltitle{\textsc{DeepSciVerify}: Verifying Scientific Claim--Citation Alignment via LLM-Driven Evidence Escalation}

\icmlsetsymbol{equal}{*}

\begin{icmlauthorlist}
\icmlauthor{Shaghayegh Sadeghi*}{yyy}
\icmlauthor{Khashayar Khajavi*}{yyy,y}
\icmlauthor{Rise Adhikari}{yyy}
\icmlauthor{Alexander Tessier}{yyy}
\end{icmlauthorlist}

\icmlaffiliation{yyy}{FirstPrinciples}
\icmlaffiliation{y}{School of Computing Science, Simon Fraser University}
\icmlcorrespondingauthor{FirstPrinciples}{research@firstprinciples.org}

\icmlkeywords{Citation verification, Claim verification, Large language models, Evidence retrieval, Evidence retrieval}

\vskip 0.3in
]

\printAffiliationsAndNotice{* Equal contribution.}


\begin{abstract}
    Misalignment between claims and their cited evidence is a common failure mode in reports generated by large language models, limiting their reliability in scientific and other high-stakes settings. We present \textsc{DeepSciVerify}, a two-stage pipeline for scientific claim--citation verification that combines abstract-level reasoning with selective escalation to passage-level evidence. The system first verifies claims using the abstract and defers uncertain cases, retrieving and analyzing full-text passages only when necessary. This design leverages complementary behaviors across LLMs, as some models are more conservative while others are more decisive under uncertainty. On the \textsc{SCitance} benchmark, \textsc{DeepSciVerify} achieves 86.7 Micro-F1, outperforming strong abstract-only baselines by +4.5 points while resolving 67\% of instances without full-text retrieval. These results suggest that selective evidence escalation improves both accuracy and efficiency in claim--citation verification. 
\end{abstract}

\section{Introduction}


Scientific claim verification \cite{wadden2022scifact, fang2025automatic, alvarez2024zero, wang2025sciver} is the task of determining whether a scientific claim is supported, contradicted, or not sufficiently resolved by evidence from the research literature. In generated scientific reports, scientific claim verification can be framed as claim--citation alignment: given a claim and its associated citation, the goal is to verify whether the cited work actually provides evidence for the claim. This distinction is important because a citation may be bibliographically valid while still being semantically misaligned with the statement it is used to support. As the volume of scientific publications continues to grow, and as AI-generated text becomes increasingly common in research workflows~\cite{liang2024monitoring,liang2024mapping,khalifa2024using,kobak2025delving}, automatic verification of claim--citation alignment has become an important requirement for maintaining the reliability and trustworthiness of scientific writing.

\begin{figure*}[t]
    \centering
    \includegraphics[width=0.9\textwidth]{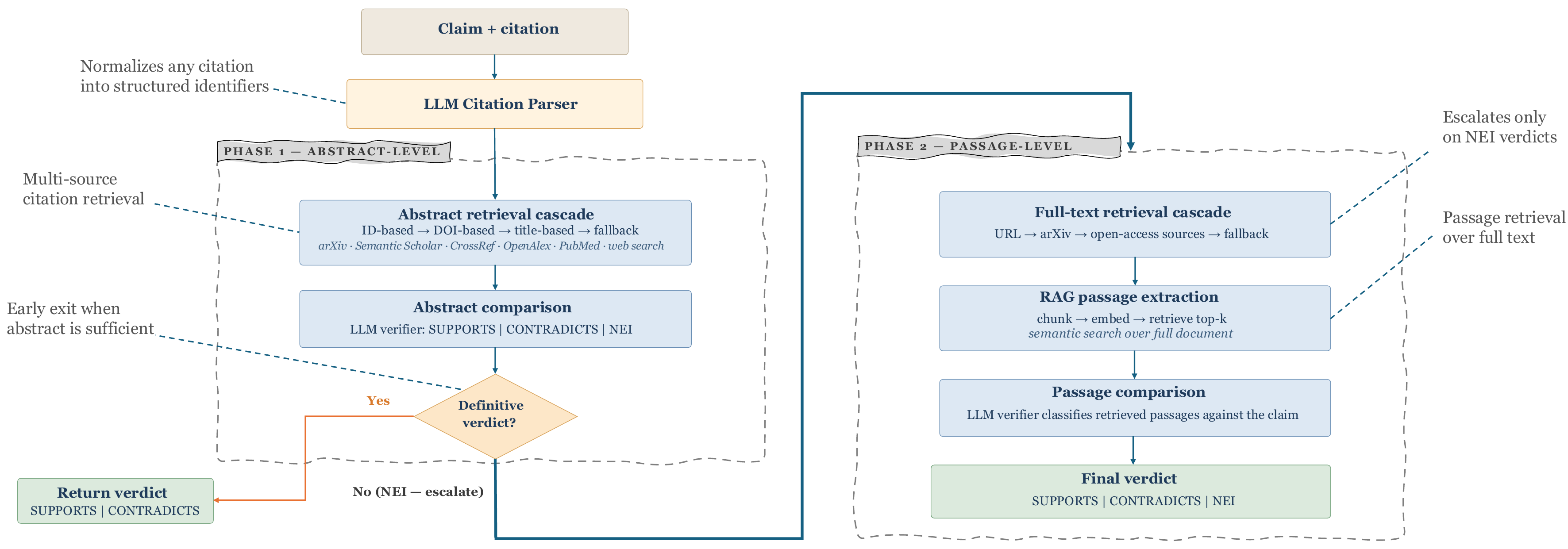}
    \caption{Overview of the proposed two-stage claim verification pipeline. The system first performs abstract-level verification with an early-exit mechanism and escalates to passage-level retrieval and verification only when abstract evidence is insufficient.}
    \label{fig:framework}
\end{figure*}


Existing work in scientific claim verification has largely been developed around benchmark settings where claims are verified against abstracts from a fixed scientific corpus. Systems such as SciFact \cite{wadden2022scifact} and SCIVER \cite{wang2025sciver} provide a controlled setting for evaluating evidence retrieval and veracity prediction, but they simplify several challenges that arise in practical claim--citation verification. In particular, abstracts provide only a compressed summary of a paper and often omit methodological details, experimental conditions, quantitative results, ablation findings, or limitations that are necessary to assess whether a cited work truly supports a claim. As a result, an abstract may be insufficient even when the cited paper itself contains relevant evidence. 

To address this limitation, in this work, we propose \textsc{DeepSciVerify}, a two-stage claim--citation verification pipeline (Figure~\ref{fig:framework}). Given a claim and its associated citation, the system first parses the citation into structured retrieval signals, retrieves the cited paper's abstract through a multi-source retrieval cascade, and performs abstract-level verification with an LLM. If the abstract provides sufficient evidence to support or contradict the claim, the system returns the verdict through an early-exit mechanism. Otherwise, when the abstract evidence is insufficient, \textsc{DeepSciVerify} escalates to full-text verification: it retrieves the complete paper when available, extracts relevant passages using RAG-based passage retrieval, and produces a final verdict from passage-level evidence.

We evaluate \textsc{DeepSciVerify} on \textsc{SCitance}\cite{alvarez2024zero}, a scientific claim verification dataset derived from citation sentences and evidence abstracts. In the three-class setting, \textsc{DeepSciVerify} achieves 86.7 Micro-F1 and 81.5 Macro-F1, outperforming the strongest abstract-only baseline by 4.5 Micro-F1 and improving over the best prior result on \textsc{SCitance} by 6.6 Micro-F1 (Section~\ref{sec:main_results}). Beyond overall performance, we analyze how different LLMs behave under evidence uncertainty. Our calibration analysis reveals distinct model-level biases: GPT-5.4 is overly cautious and frequently defaults to \textsc{nei}, while Claude Sonnet 4.6 is more overconfident and tends to commit to \textsc{supports} or \textsc{contradicts} even when the available evidence is incomplete (Section~\ref{sec:calibration}). Finally, our escalation analysis shows that \textsc{DeepSciVerify} resolves 67.0\% of instances at the abstract level and escalates only the remaining 33.0\% to full-text verification, yielding a net gain on it's correct predictions(Section~\ref{sec:escalation}).


Our work makes the following contributions:
\begin{itemize}[leftmargin=*, itemsep=2pt, topsep=2pt]
    \item We propose \textsc{DeepSciVerify}, a two-stage claim--citation verification pipeline that combines abstract-level early exit with selective escalation to passage-level evidence.
    \item We develop multi-source retrieval cascades for citation parsing, abstract retrieval, and full-text retrieval, enabling verification under incomplete or noisy citation metadata.
    \item We analyze model-level calibration in abstract-only verification and show that different LLMs exhibit distinct biases under evidence uncertainty, motivating our use of complementary models across pipeline stages.
    \item We show that selective evidence escalation improves over strong abstract-only baselines on \textsc{SCitance}, and provide ablations demonstrating the importance of RAG-based passage extraction \footnote{Code and data are available upon request.}.
\end{itemize}


\section{Related Work}

Scientific claim verification is a subarea of fact verification focused on determining whether scientific evidence supports, contradicts, or is insufficient to verify a claim. FEVER \cite{thorne2018fever} established the general evidence-based verification formulation, while SciFact \cite{wadden2020fact,wadden2022scifact} adapted it to the scientific domain by pairing expert-written claims with evidence abstracts. Recent benchmarks further extend this setting: \textsc{SCitance} \cite{alvarez2024zero} uses citation sentences as claims, and SCIVER \cite{wang2025sciver} evaluates verification in multimodal scientific contexts.

Our work is also related to retrieval-augmented generation and dense retrieval. RAG \cite{lewis2020rag}, dense passage retrieval \cite{karpukhin2020dense}, and sentence embedding methods \cite{reimers2019sentence} show that retrieval can improve reasoning on knowledge-intensive tasks and enable scalable search over long documents. These methods are especially relevant for scientific verification, where necessary evidence may appear outside the abstract.

Despite these advances, most scientific claim verification systems remain centered on abstract-level evidence. In practical claim--citation verification, the cited paper may need to be resolved from multiple sources, and the relevant evidence may appear only in the full text. \textsc{DeepSciVerify} addresses this gap by combining abstract-level verification with selective escalation to passage-level evidence.






\section{Problem Definition}
\label{sec:problem_definition}

Let $c$ denote a scientific claim and $r$ a cited reference. The goal is to determine a veracity label $y \in \{\texttt{SUPPORTS}, \texttt{CONTRADICTS}, \texttt{NEI}\}$ for the pair $(c, r)$ based on evidence from the cited work. Here, \texttt{NEI} denotes \textit{not enough information}, following the label schema used in prior work \cite{wadden2020fact, thorne2018fever}. This formulation captures the task of claim--citation alignment, where the objective is to verify whether the cited reference provides sufficient evidence to support or contradict the claim, or whether the available evidence is inconclusive.

\section{Methodology}
\label{sec:methodology}

We present \textsc{DeepSciVerify}, a two-stage pipeline for claim--citation verification (Figure~\ref{fig:framework}). The system first attempts to verify claims using abstract-level evidence and escalates to passage-level analysis only when necessary. Each stage consists of a retrieval step followed by an evidence comparison step.

The pipeline operates as a staged decision process. In the first stage, an abstract-level verifier $f_a$ produces a prediction $\hat{y}_a$ based on abstract evidence $e_a$. If $\hat{y}_a \neq \texttt{NEI}$, the system returns this prediction. Otherwise, the system retrieves passage-level evidence $e_p = \{p_1, \ldots, p_k\}$ from the full text and applies a second verifier $f_p$ to produce the final label:
\begin{equation}
\label{eq:verdict}
\hat{y} =
\begin{cases}
f_a(c, e_a), & \text{if } f_a(c, e_a) \neq \texttt{NEI}, \\[4pt]
f_p(c, e_p), & \text{otherwise.}
\end{cases}
\end{equation}

\subsection{Phase 1: Abstract-Level Verification}
\label{sec:phase1}

Phase~1 verifies the claim using only the abstract of the cited paper, resolving a substantial proportion of instances without full-text access \cite{wadden2020fact, wadden2021sciver}.





\subsubsection{Abstract Retrieval}
\label{sec:abstract_retrieval}

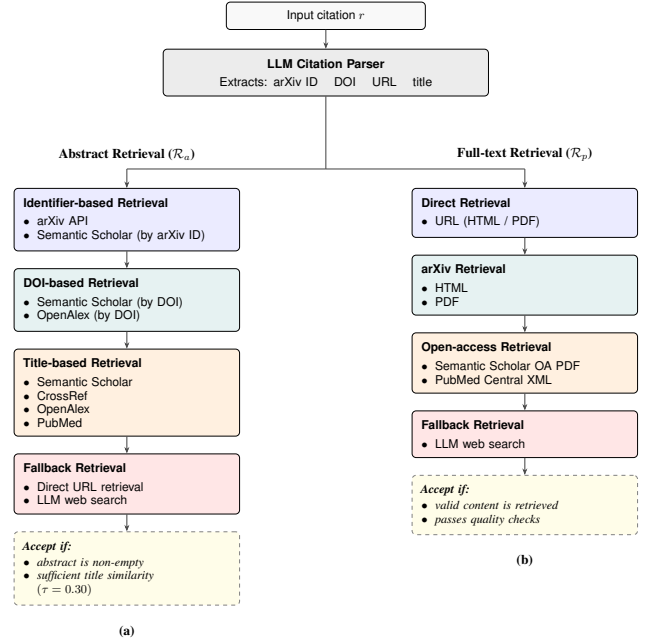
\begin{figure}[t]
\centering
\resizebox{\linewidth}{!}{%
\begin{tikzpicture}[
  font=\small\sffamily,
  node distance=4.5mm,
  group/.style={
    draw, rounded corners=3pt, inner sep=7pt,
    minimum width=58mm, align=left, text width=54mm,
  },
  parser/.style={
    draw, rounded corners=3pt, inner sep=7pt,
    fill=gray!15, minimum width=85mm, align=center, text width=78mm,
  },
  input/.style={
    draw, rounded corners=3pt, inner sep=6pt,
    fill=gray!5, minimum width=52mm, align=center,
  },
  gate/.style={
    draw=black!50, dashed, rounded corners=3pt,
    inner sep=7pt, fill=yellow!10, align=left, text width=54mm,
    font=\small\itshape,
  },
  arr/.style={-{Stealth[length=5pt]}, thick, draw=black!70},
  line/.style={thick, draw=black!70},
  title/.style={font=\normalsize\bfseries, align=center},
]
\node[input] (input) {Input citation $r$};
\node[parser, below=5mm of input] (parser) {%
  \textbf{LLM Citation Parser}\\[4pt]
  Extracts: arXiv ID \quad DOI \quad URL \quad title
};
\draw[arr] (input) -- (parser);
\node[group, fill=blue!8, below=24mm of parser, xshift=-52mm] (a1) {%
  \textbf{Identifier-based Retrieval}\\[4pt]
  $\bullet$\; arXiv API\\
  $\bullet$\; Semantic Scholar (by arXiv ID)
};
\node[title, above=6mm of a1]
  {Abstract Retrieval ($\mathcal{R}_a$)};
\node[group, below=of a1, fill=teal!10] (a2) {%
  \textbf{DOI-based Retrieval}\\[4pt]
  $\bullet$\; Semantic Scholar (by DOI)\\
  $\bullet$\; OpenAlex (by DOI)
};
\node[group, below=of a2, fill=orange!12] (a3) {%
  \textbf{Title-based Retrieval}\\[4pt]
  $\bullet$\; Semantic Scholar\\
  $\bullet$\; CrossRef\\
  $\bullet$\; OpenAlex\\
  $\bullet$\; PubMed
};
\node[group, below=of a3, fill=red!10] (a4) {%
  \textbf{Fallback Retrieval}\\[4pt]
  $\bullet$\; Direct URL retrieval\\
  $\bullet$\; LLM web search
};
\node[gate, below=of a4] (ag) {%
  \textbf{Accept if:}\\[2pt]
  $\bullet$\; abstract is non-empty\\
  $\bullet$\; sufficient title similarity\\
  \hspace*{1.1em}$(\tau = 0.30)$
};
\node[group, fill=blue!8, below=24mm of parser, xshift=52mm] (b1) {%
  \textbf{Direct Retrieval}\\[4pt]
  $\bullet$\; URL (HTML / PDF)
};
\node[title, above=6mm of b1]
  {Full-text Retrieval ($\mathcal{R}_p$)};
\node[group, below=of b1, fill=teal!10] (b2) {%
  \textbf{arXiv Retrieval}\\[4pt]
  $\bullet$\; HTML\\
  $\bullet$\; PDF
};
\node[group, below=of b2, fill=orange!12] (b3) {%
  \textbf{Open-access Retrieval}\\[4pt]
  $\bullet$\; Semantic Scholar OA PDF\\
  $\bullet$\; PubMed Central XML
};
\node[group, below=of b3, fill=red!10] (b4) {%
  \textbf{Fallback Retrieval}\\[4pt]
  $\bullet$\; LLM web search
};
\node[gate, below=of b4] (bg) {%
  \textbf{Accept if:}\\[2pt]
  $\bullet$\; valid content is retrieved\\
  $\bullet$\; passes quality checks
};
\node[coordinate, below=19mm of parser] (split) {};
\draw[line] (parser.south) -- (split);
\draw[line] (split) -- (split -| a1.north);
\draw[arr]  (split -| a1.north) -- (a1.north);
\draw[line] (split) -- (split -| b1.north);
\draw[arr]  (split -| b1.north) -- (b1.north);
\foreach \i/\j in {a1/a2,a2/a3,a3/a4,a4/ag}{\draw[arr] (\i) -- (\j);}
\foreach \i/\j in {b1/b2,b2/b3,b3/b4,b4/bg}{\draw[arr] (\i) -- (\j);}
\node[font=\normalsize\bfseries, below=4mm of ag] {(a)};
\node[font=\normalsize\bfseries, below=4mm of bg] {(b)};
\end{tikzpicture}}
\caption{Overview of the two-stage retrieval architecture in. An input citation $r$ is first processed by an LLM-based parser to extract structured signals (e.g., arXiv ID, DOI, URL, and title), which guide both retrieval cascades. (a) Abstract-level retrieval ($\mathcal{R}_a$) operates as a hierarchical cascade from identifier-based lookup to fallback strategies. (b) Full-text retrieval ($\mathcal{R}_p$) retrieves and processes complete documents when abstract-level evidence is insufficient. Each cascade halts at the first stage that satisfies its acceptance criterion, and the retrieved evidence is forwarded to the corresponding verification module.}
\label{fig:retrieval-cascades}
\end{figure}

The abstract retrieval stage operates as a cascade guided by signals extracted from the citation, as shown in Figure~\ref{fig:retrieval-cascades}(a). Given a claim--citation pair $(c, r)$, the goal is to retrieve an abstract $e_a$ corresponding to the cited reference $r$.

\paragraph{Citation Parsing.}
The process begins with an LLM-based citation parser, which converts the raw citation string into structured fields, including arXiv ID, DOI, URL, and title. These fields serve as retrieval signals that determine which branches of the cascade can be activated. By standardizing heterogeneous citation formats into a consistent representation, the parser enables reliable downstream retrieval.

\paragraph{Identifier-based Retrieval.}
When an arXiv identifier is available, the system performs direct lookup using the arXiv API, followed by resolution through Semantic Scholar~\cite{kinney2023semantic}. This stage provides the most precise retrieval signal, as identifiers uniquely specify a document.

\paragraph{DOI-based Retrieval.}
If a DOI is present, the system queries multiple scholarly APIs, including Semantic Scholar~\cite{kinney2023semantic} and OpenAlex~\cite{priem2022openalex}. DOI-based lookup offers high coverage and reliability across publishers and disciplines.

\paragraph{Title-based Retrieval.}
In the absence of reliable identifiers, the system falls back to title-based search across multiple sources, including Semantic Scholar, CrossRef~\cite{hendricks2020crossref}, OpenAlex, and PubMed~\cite{sayers2022pubmed}. This stage trades precision for recall, allowing recovery of references with incomplete or noisy metadata.

\paragraph{Fallback Retrieval.}
If all structured retrieval attempts fail, the system applies fallback strategies, including direct URL-based retrieval and LLM-assisted web search. This stage is designed to maximize coverage in cases where bibliographic metadata is missing or corrupted. Prompts for LLM-assisted abstract search are provided in Appendix~\ref{app:abstract_search_prompt}.

\paragraph{Acceptance Criterion.}
At each stage, retrieved candidates are validated using a lightweight acceptance gate. An abstract is accepted if it is non-empty and sufficiently matches the citation, measured via title similarity. The cascade terminates at the first stage that produces a valid result, ensuring both efficiency and robustness. Additional implementation details, including acceptance criteria, are provided in Appendix~\ref{app:retrieval_details}.

\subsubsection{Abstract Comparison and Early Exit}
\label{sec:abstract_comparison}

Once the abstract $e_a$ is retrieved, the system performs claim--evidence classification using a large language model. The model receives the claim $c$ and the abstract $e_a$ as input and predicts a label:
\begin{equation}
\hat{y}_a = f_a(c, e_a), \quad \hat{y}_a \in \{\texttt{SUPPORTS}, \texttt{CONTRADICTS}, \texttt{NEI}\}.
\end{equation}

The model is prompted to assess whether the abstract provides sufficient evidence to support or contradict the claim, or whether the available evidence is inconclusive. Full prompts are provided in Appendix~\ref{app:prompts}.

The predicted label $\hat{y}_a$ also serves as a decision signal for the pipeline. If $\hat{y}_a \in \{\texttt{SUPPORTS}, \texttt{CONTRADICTS}\}$, the system returns this prediction directly, avoiding unnecessary full-text retrieval. Otherwise, if $\hat{y}_a = \texttt{NEI}$, the pipeline escalates to the full-text retrieval stage. This design allocates computational resources according to instance difficulty \cite{wadden2022scifact}.

\subsection{Phase 2: Passage-Level Verification}
\label{sec:phase2}

Phase~2 is triggered when $\hat{y}_a = \texttt{NEI}$. In this case, abstract-level evidence is insufficient, and the system performs deeper verification using full-text content.

\subsubsection{Full-Text Retrieval}
\label{sec:fulltext_retrieval}

The full-text retrieval stage follows a structured cascade similar to the abstract retrieval process, as illustrated in Figure~\ref{fig:retrieval-cascades}(b). Given a cited reference $r$, the goal is to retrieve the full document from which fine-grained evidence can be extracted.

\paragraph{Direct Retrieval.}
The system first attempts to retrieve content directly from the URL associated with the citation, when available. This includes both HTML pages and PDF documents, which are parsed into raw text.

\paragraph{arXiv Retrieval.}
If an arXiv identifier is available, the system retrieves the document from arXiv. When possible, the HTML version is preferred due to its cleaner structure, with PDF used as a fallback.

\paragraph{Open-access Retrieval.}
Next, the system queries open-access repositories, including Semantic Scholar~\cite{kinney2023semantic} and PubMed Central (PMC)~\cite{sayers2022pubmed}, to obtain full-text documents in PDF or structured XML formats.

\paragraph{Fallback Retrieval.}
If all previous stages fail, the system performs LLM-assisted web search using the cited paper title. The search prompt asks the model to return a freely accessible direct full-text URL in PDF, XML, or HTML format, prioritizing open repositories such as PMC, arXiv, preprint servers, and institutional repositories. To reduce hallucinated retrievals, the model is instructed to return \texttt{found: false} rather than guess or invent a URL when no direct full-text link is found. Prompts for LLM-assisted full-text search are provided in Appendix~\ref{app:fulltext_search_prompt}.

\paragraph{Acceptance Criterion.}
Retrieved documents are validated using a lightweight acceptance criterion. A document is accepted if it contains sufficient textual content and passes basic quality checks (e.g., excluding corrections or errata). As in Phase~1, the cascade terminates at the first stage that produces a valid result. Additional implementation details, including filtering criteria, are provided in Appendix~\ref{app:retrieval_details}.





\subsubsection{Passage Extraction}
\label{sec:passage_extraction}

Once the full text is retrieved, the system extracts a set of relevant passages $e_p = \{p_1, p_2, \ldots, p_k\}$ that are most likely to contain evidence for the claim $c$. Our approach follows the retrieval-augmented generation (RAG) paradigm \cite{lewis2020rag}. The full text is segmented into chunks, each of which is mapped to a dense vector representation using a sentence embedding model \cite{reimers2019sentence}. The claim is embedded in the same space, and the top-$k$ most relevant passages are selected based on cosine similarity:
\begin{equation}
e_p = \text{top-}k \left( \frac{\mathbf{v}_c \cdot \mathbf{v}_{p_i}}{\|\mathbf{v}_c\| \, \|\mathbf{v}_{p_i}\|} \right),
\end{equation}
where $\mathbf{v}_c$ denotes the embedding of the claim and $\mathbf{v}_{p_i}$ denotes the embedding of the $i$-th passage. This approach scales efficiently to long documents and enables focused retrieval of evidence without requiring the entire document to be processed by the LLM.

Furthermore, we also consider an LLM-based extraction strategy in which the full text and claim are provided directly to a model, which selects relevant passages based on semantic reasoning. This approach is evaluated as an ablation (Section~\ref{sec:ablation}) to compare against the RAG-based method.

\subsubsection{Passage Comparison}
\label{sec:passage_comparison}

The extracted passages $e_p$ are then compared against the claim $c$ using an LLM-based classifier:
\begin{equation}
\hat{y}_p = f_p(c, e_p), \quad \hat{y}_p \in \{\texttt{SUPPORTS}, \texttt{CONTRADICTS}, \texttt{NEI}\}.
\end{equation}

The classifier evaluates the passages collectively and determines whether they provide sufficient evidence to support or contradict the claim. If the evidence remains inconclusive, the system outputs \texttt{NEI}.

\subsection{Final Verdict}
\label{sec:final_verdict}

The overall verdict $\hat{y}$ follows Equation~\ref{eq:verdict}: the system returns the Phase~1 prediction if definitive, or the Phase~2 prediction otherwise. When evidence remains inconclusive after both phases, the system outputs \texttt{NEI}, prioritizing precision over unsupported conclusions.

\section{Experiments}

\subsection{Dataset}
\label{sec:dataset}

We evaluate on \textsc{SCitance} \cite{alvarez2024zero}, a scientific claim verification dataset derived from \textsc{SciFact} \cite{wadden2020fact}. Unlike \textsc{SciFact}, which uses expert-written claims, \textsc{SCitance} uses citation sentences (\textit{citances}) as claims, making it closely aligned with claim--citation verification. Each citance is paired with an evidence abstract and labeled as \texttt{SUPPORTS}, \texttt{CONTRADICTS}, or \texttt{NEI}. Contradiction examples were generated by negating citances with GPT-3.5 \cite{alvarez2024zero}. The dataset contains 656 instances in total: 251 \texttt{SUPPORTS}, 225 \texttt{CONTRADICTS}, and 180 \texttt{NEI}, split into 467 train, 98 dev, and 91 test instances.

Following our experimental design, we use the combined train and development splits to analyze model behavior and calibration, and reserve the test split exclusively for final evaluation and comparison against baselines.




\subsection{Experimental Setting}
\label{sec:experimental_setting}

We evaluate both two-class and three-class settings. In the two-class setting, the model predicts only \texttt{SUPPORTS} or \texttt{CONTRADICTS}, whereas in the three-class setting it additionally predicts \texttt{NEI}. We report Micro-F1 and Macro-F1 in both settings, and in the three-class setting, we also report the binary support versus not-support metric used in prior work. We compare \textsc{DeepSciVerify} against several strong LLM baselines, including GPT-4, GPT-5.4, Gemini 2.5 Flash, and Claude Sonnet 4.6. In this work, All LLM-driven components are accessed via hosted provider APIs. All baseline models are evaluated in a zero-shot setting, with temperature set to 0.

For \textsc{DeepSciVerify}, we use the two-stage pipeline described in Section~\ref{sec:methodology}, consisting of abstract-level verification followed by optional full-text retrieval and passage-level verification when the abstract is insufficient. The specific design choices of the pipeline, including the selection of LLMs and retrieval configurations, are informed by the model behavior and calibration analysis presented in Section~\ref{sec:calibration}. Final performance is reported on the held-out test set (Section~\ref{sec:main_results}), and ablation studies examining key components of the system are provided in Section~\ref{sec:ablation}.


\paragraph{Implementation details.}
For passage-level retrieval, the full text is segmented into chunks of 5{,}000 tokens. We retrieve the top-2 passages based on cosine similarity with a threshold of 0.5. The embedding model used is \texttt{text-embedding-3-small}~\cite{openai2024embedding}. The choice of models and retrieval parameters is guided by the analysis in Section~\ref{sec:calibration}. All prompts are provided in Appendix~\ref{app:prompts}, and full model and experiment configuration details are reported in Appendix~\ref{app:model_config}.

\section{Results}

\subsection{Retrieval Coverage Analysis}
\label{sec:retrieval_analysis}

To evaluate the effectiveness of our retrieval modules (Sections~\ref{sec:abstract_retrieval} and \ref{sec:fulltext_retrieval}), we analyze their coverage over all unique cited papers in \textsc{SCitance}. We extract 412 unique references across the train, development, and test splits, and attempt to retrieve both abstracts and full text for each paper.

Table~\ref{tab:retrieval_coverage} summarizes the results. Abstract retrieval achieves high coverage, successfully retrieving abstracts for 391 papers (94.9\%). Full-text retrieval is more challenging but still recovers complete documents for 333 papers (80.8\%). In 74 cases (18.0\%), the abstract is available but full text is not retrieved, while only 5 papers (1.2\%) have no retrieved evidence.

These results highlight two important aspects of the task. First, while abstracts are widely available, full-text access is not guaranteed, reinforcing the need for a system that can operate effectively under partial evidence. Second, the retrieval pipeline provides full-text access for the majority of cited papers, enabling passage-level verification in many cases where abstract evidence may be insufficient. Additional source-level, split-level, and latency analyses are provided in Appendix~\ref{app:retrieval_coverage_latency}.

\begin{table}[t]
\centering
\small
\caption{Coverage of abstract and full-text retrieval over 412 unique papers in \textsc{SCitance}.}
\label{tab:retrieval_coverage}

\begin{tabularx}{\columnwidth}{>{\raggedright\arraybackslash}Xcc}
\toprule
\textbf{Retrieval Type} & \textbf{Count} & \textbf{Percentage} \\
\midrule
Abstract retrieved  & 391 & 94.9\% \\
Full text retrieved & 333 & 80.8\% \\
Abstract only       & 74  & 18.0\% \\
No retrieval        & 5   & 1.2\% \\
\bottomrule
\end{tabularx}

\end{table}

\subsection{Calibration Analysis}
\label{sec:calibration}

\begin{figure*}[t]
    \centering
    \includegraphics[width=\textwidth]{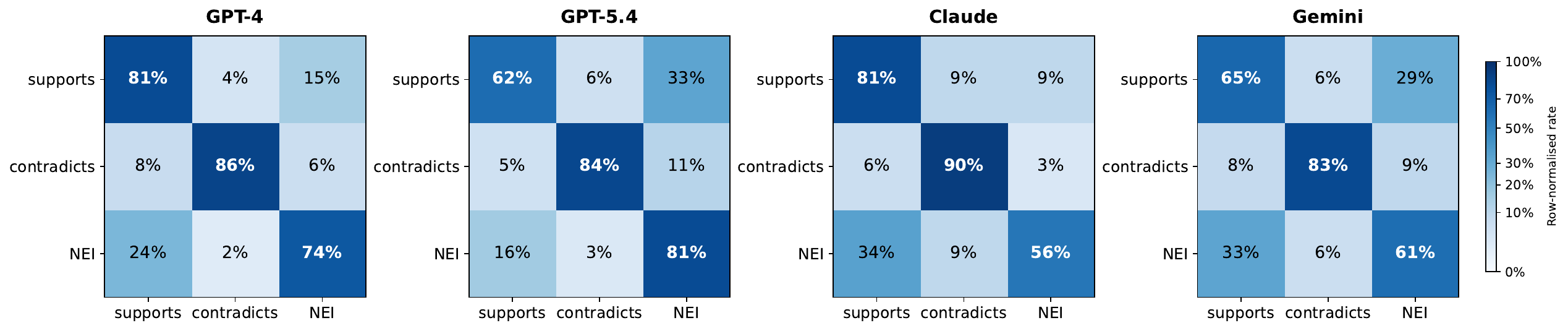}
    \caption{Row-normalized confusion matrices for abstract-only verification on the \textsc{SCitance} train+dev set. Each model is given the claim and evidence abstract and predicts \texttt{SUPPORTS}, \texttt{CONTRADICTS}, or \texttt{NEI}. Rows indicate gold labels and columns indicate predicted labels. The matrices reveal distinct calibration patterns across models, including conservative behavior, overcommitment, and class-specific error asymmetries. We compute these matrices on the combined train+dev split ($n = 565$) to obtain more stable calibration estimates; see Section~\ref{sec:calibration}.}
    \label{fig:confusion_matrices}
\end{figure*}

\begin{figure}
    \centering
    \includegraphics[width=0.9\columnwidth]{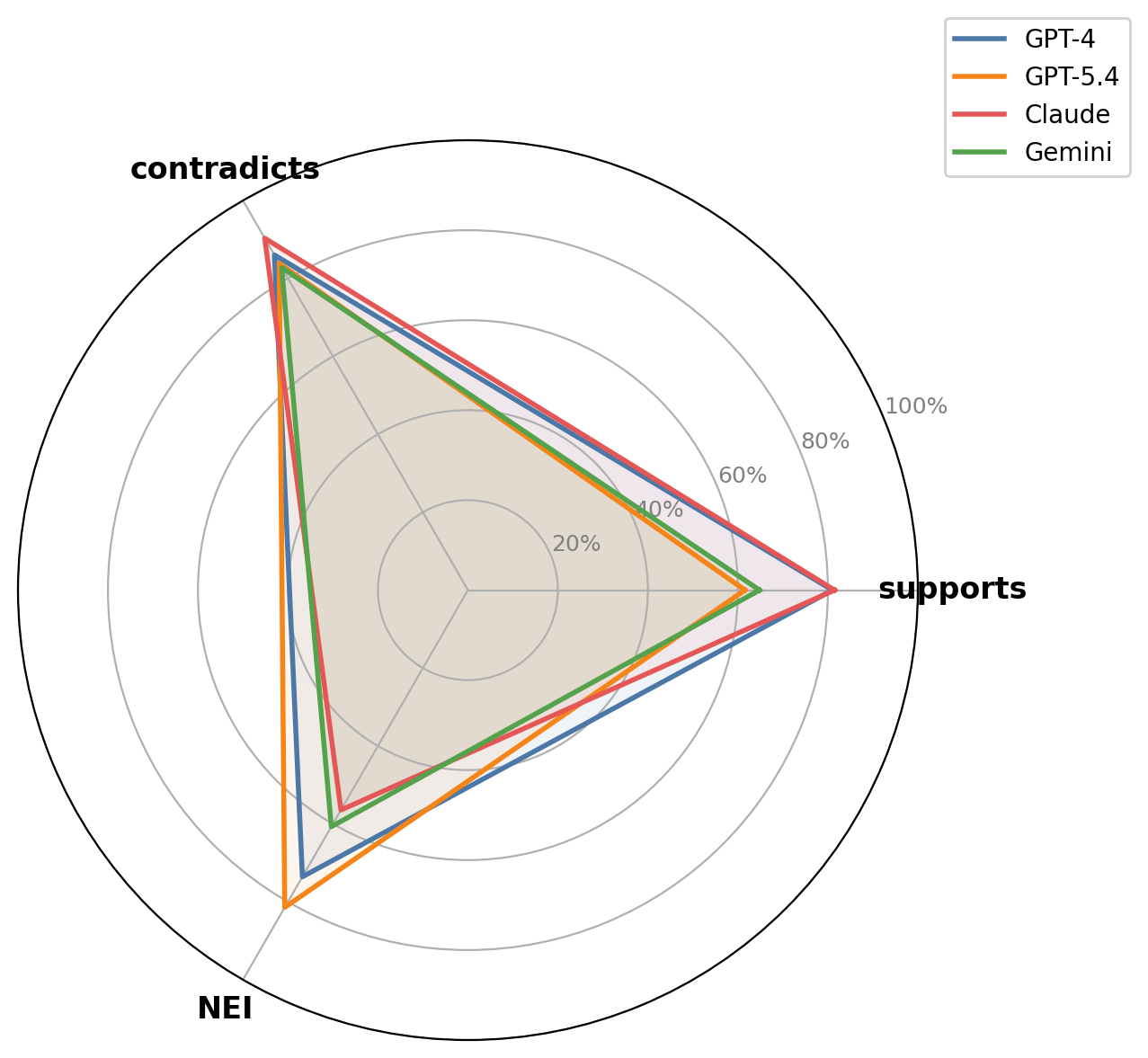}
    \caption{Per-class recall comparison across models on the \textsc{SCitance} train+dev set, highlighting differences in calibration behavior.}
    \label{fig:calibration_radar}
\end{figure}

We first analyze how different LLMs behave in the abstract-level verification setting on the combined train+dev split ($n = 565$). In these experiments, each model is given the claim together with the corresponding evidence abstract and asked to predict one of \texttt{SUPPORTS}, \texttt{CONTRADICTS}, or \texttt{NEI}. Our goal is to determine whether the models exhibit similar behavior or whether they have distinct implicit biases under the same evidence conditions.  Figure~\ref{fig:confusion_matrices} provides a detailed view of error distributions across labels, while Figure~\ref{fig:calibration_radar} summarizes per-class recall in a compact form. Together, these visualizations show that the models differ substantially not only in overall performance, but also in how they allocate predictions across the three labels.

\begin{table*}[t]
\centering
\small
\caption{Qualitative calibration analysis of abstract-only LLM verification on the \textsc{SCitance} train+dev set ($n = 565$). The analysis is based on the row-normalized confusion matrices in Figure~\ref{fig:confusion_matrices}.}
\label{tab:calibration}
\label{tab:bias}
\begin{tabular}{p{0.20\textwidth} p{0.74\textwidth}}
\toprule
\textbf{Model} & \textbf{Observed Calibration Pattern} \\
\midrule
GPT-4 &
\textcolor{blue}{\textbf{Balanced.}} Shows the most even behavior across classes, with recall of 81\% for \texttt{SUPPORTS}, 86\% for \texttt{CONTRADICTS}, and 74\% for \texttt{NEI}. This makes it the most stable verifier across the three-way decision. \\[0.5em]

GPT-5.4 &
\textcolor{blue}{\textbf{Conservative.}} Achieves strong \texttt{NEI} recall (81\%) but lower \texttt{SUPPORTS} recall (62\%), with 33\% of true \texttt{SUPPORTS} cases predicted as \texttt{NEI}. This makes it well suited for an early-stage verifier that should escalate uncertain cases rather than overcommit. \\[0.5em]

Gemini 2.5 Flash &
\textcolor{blue}{\textbf{Cautious with moderate imbalance.}} Like GPT-5.4, it often maps true \texttt{SUPPORTS} cases to \texttt{NEI} (29\%), while achieving 83\% recall on \texttt{CONTRADICTS} and 61\% on \texttt{NEI}. \\[0.5em]

Claude Sonnet 4.6 &
\textcolor{blue}{\textbf{Overconfident.} }Performs strongly on decisive labels, with 81\% \texttt{SUPPORTS} recall and 90\% \texttt{CONTRADICTS} recall, but has weaker \texttt{NEI} recall (56\%). It frequently converts true \texttt{NEI} cases into \texttt{SUPPORTS} or \texttt{CONTRADICTS}, indicating a tendency to overcommit when abstract evidence is incomplete. \\
\bottomrule
\end{tabular}
\end{table*}

Surprisingly, GPT-4 emerges as the most well-calibrated abstract-level verifier, exhibiting consistently strong and balanced recall across all three classes (81\% \texttt{SUPPORTS}, 86\% \texttt{CONTRADICTS}, 74\% \texttt{NEI}; Figure~\ref{fig:confusion_matrices}). Table~\ref{tab:calibration} summarizes these patterns qualitatively and connects each model's error profile to its role in our pipeline design. This balanced behavior is also reflected in Figure~\ref{fig:calibration_radar}, where GPT-4 exhibits the most uniform recall profile across all three classes. In contrast, GPT-5.4 which is a newer model, exhibits a markedly conservative bias: while it achieves the highest \texttt{NEI} recall (81\%), it does so at the expense of \texttt{SUPPORTS} recall (62\%), misclassifying 33\% of true \texttt{SUPPORTS} instances as \texttt{NEI}. This asymmetry suggests that GPT-5.4 systematically defers judgment when abstract evidence is even mildly incomplete. While detrimental in a single-stage setting, this behavior is desirable in a staged pipeline, where \texttt{NEI} serves as a trigger for escalation rather than a terminal prediction.

Claude Sonnet 4.6 displays the opposite calibration profile. It is highly decisive, achieving strong recall on \texttt{SUPPORTS} (81\%) and the highest recall on \texttt{CONTRADICTS} (90\%), but at the cost of substantially weaker \texttt{NEI} detection (56\%). As shown in Figure~\ref{fig:confusion_matrices}, 34\% of true \texttt{NEI} instances are misclassified as \texttt{SUPPORTS}, indicating a strong tendency to overcommit under uncertainty. This asymmetry is visually apparent in Figure~\ref{fig:calibration_radar}, where Claude's performance is skewed toward decisive labels at the expense of \texttt{NEI}. Gemini 2.5 Flash exhibits a similar but less extreme pattern, combining moderate \texttt{CONTRADICTS} recall (83\%) with elevated confusion between \texttt{SUPPORTS} and \texttt{NEI} (29\% of \texttt{SUPPORTS} mapped to \texttt{NEI}). Taken together, these results reveal that LLMs do not simply differ in overall accuracy, but exhibit distinct and systematic calibration biases along the caution--overconfidence spectrum. This variation is central to our design: rather than selecting a single ``best'' model, \textsc{DeepSciVerify} leverages these complementary behaviors by assigning conservative models to early-stage filtering and more balanced models to final decision-making.

These calibration differences motivate our pipeline design. We use GPT-5.4 as the abstract-level verifier because its conservative behavior makes it suitable for deciding when to escalate beyond the abstract. For passage-level verification, where the system has access to more focused evidence, we use GPT-4 because of its more balanced behavior across all three labels. This combination allows \textsc{DeepSciVerify} to avoid overcommitting at the abstract stage while still making calibrated final decisions once passage-level evidence is available.

\subsection{Main Results}
\label{sec:main_results}

Table~\ref{tab:results} presents the main results on the \textsc{SCitance} test set. Since \textsc{SCitance} already provides evidence abstracts, all abstract-only baselines use the provided abstract as input. In the full-text setting, we additionally equip the model with the full-text retriever described in Section~\ref{sec:fulltext_retrieval}, allowing it to retrieve and use passage-level evidence when the abstract-level prediction is \texttt{NEI}.

In the two-class setting without \texttt{NEI}, all models perform strongly, with Claude Sonnet 4.6 achieving the highest performance (93.2 Micro-F1). This suggests that modern LLMs are effective when forced to decide only between support and contradiction. However, this setting removes the need to detect insufficient evidence, which is central to realistic scientific verification.

In the full three-class setting, performance drops across all abstract-only baselines, showing that correctly identifying \texttt{NEI} remains challenging. Providing access to full-text evidence improves most models, indicating that information beyond the abstract can be useful. However, the gains from naive full-text usage are limited: GPT-4 decreases from 82.2 to 81.3 Micro-F1, while Claude Sonnet 4.6 improves only marginally from 81.1 to 82.4.

\textsc{DeepSciVerify} achieves the best overall performance, reaching 86.7 Micro-F1 and 81.5 Macro-F1. This outperforms the strongest abstract-only baseline by +4.5 Micro-F1 and improves over the best number reported by \citet{alvarez2024zero} on \textsc{SCitance} by +6.6 Micro-F1. Since \textsc{SCitance} uses a 91-instance test split, these gains should be interpreted with appropriate caution, but they are consistent across Micro-F1, Macro-F1, and Sup./Not Sup. Overall, the results suggest that full-text evidence is most effective when used selectively: \textsc{DeepSciVerify} resolves confident cases at the abstract level and escalates uncertain cases to passage-level verification, avoiding both premature decisions and unnecessary noise from full-text processing. We provide a detailed breakdown of this escalation behavior in Appendix~\ref{sec:escalation}.


\begin{table*}[t]
\centering
\small
\caption{Results on the \textsc{SCitance} test set. $\Delta$ shows absolute improvement
  over the best abstract-only baseline in each setting. Best results per setting are \textbf{bolded}.}
\label{tab:results}
\begin{tabularx}{\textwidth}{@{} l l X ccc r @{}}
\toprule
\textbf{Setting} & \textbf{Evidence} & \textbf{Model}
  & \textbf{Micro-F1} & \textbf{Macro-F1} & \textbf{Sup./Not Sup.}
  & \textbf{$\Delta$ Micro} \\
\midrule
\multirow{4}{*}{\shortstack[l]{Without NEI\\(2-class)}}
  & \multirow{4}{*}{Abstract}
  & GPT-4                         & 91.9          & 91.9          & --- & --- \\
  & & GPT-5.4                      & 87.8          & 87.8          & --- & --- \\
  & & Gemini 2.5 Flash             & 86.5          & 86.4          & --- & --- \\
  & & Claude Sonnet 4.6            & \textbf{93.2} & \textbf{93.2} & --- & --- \\
\midrule
\multirow{9}{*}{\shortstack[l]{With NEI\\(3-class)}}
  & \multirow{4}{*}{Abstract}
  & GPT-4                         & 82.2          & 78.8          & 86.7 & ref. \\
  & & GPT-5.4                      & 74.4          & 72.4          & 81.1 & \textcolor{red!70!black}{--7.8} \\
  & & Gemini 2.5 Flash             & 74.4          & 70.8          & 75.6 & \textcolor{red!70!black}{--7.8} \\
  & & Claude Sonnet 4.6            & 81.1          & 73.8          & 85.6 & \textcolor{red!70!black}{--1.1} \\
\cmidrule(lr){2-7}
  & \multirow{5}{*}{Full-text}
  & GPT-4                         & 81.3          & 76.7          & 84.6 & \textcolor{red!70!black}{--0.9} \\
  & & GPT-5.4                      & 79.1          & 76.1          & 81.3 & \textcolor{red!70!black}{--3.1} \\
  & & Gemini 2.5 Flash             & 75.8          & 71.0          & 78.0 & \textcolor{red!70!black}{--6.4} \\
  & & Claude Sonnet 4.6            & 82.4          & 76.2          & 85.7 & \textcolor{green!50!black}{+0.2} \\
\rowcolor{green!12}
  & & \textsc{DeepSciVerify} (Ours)
  & \textbf{86.7} & \textbf{81.5} & \textbf{88.9}
  & \textcolor{green!50!black}{\textbf{+4.5}} \\
\bottomrule
\end{tabularx}
\end{table*}

\subsection{Ablation Analysis}
\label{sec:ablation}

Table~\ref{tab:ablation} isolates the contribution of the passage extraction component in Phase~2. Starting from the abstract-only GPT-4 baseline (82.2 Micro-F1), adding passage-level verification consistently improves performance, confirming that full-text evidence provides additional signal beyond abstracts alone.

However, the choice of extraction strategy is critical. In the LLM-based extractor, the paper is first truncated and converted into line-numbered text; the model then receives the claim and numbered paper and returns ranked, non-overlapping line ranges rather than generating passage text directly. This constrains the extractor to select spans from the source document and reduces hallucinated passage content. Using an LLM-based extractor yields modest gains (+1.1 Micro-F1, +2.0 Macro-F1), but \emph{reduces} the Sup./Not Sup.\ metric from 86.7 to 83.3. This indicates a shift toward over-prediction of \texttt{SUPPORTS}, where some instances that should remain \texttt{NEI} are instead assigned a positive verdict. In other words, the LLM extractor improves recall by surfacing more candidate evidence, but at the cost of precision, likely because it may select passages that appear semantically relevant yet do not constitute strong supporting evidence.

In contrast, the RAG-based extractor (which we explained in Section~\ref{sec:passage_extraction}) improves all metrics simultaneously, achieving the best overall performance (+4.5 Micro-F1). By grounding passage selection in embedding-based similarity, the RAG pipeline retrieves more focused and evidence-aligned text segments, reducing spurious matches and limiting false positive support predictions. This leads to better balance across all three classes, particularly in preserving correct \texttt{NEI} decisions while still enabling corrections when sufficient evidence exists.

Overall, these results highlight that the benefit of Phase~2 does not come merely from accessing full text, but from \emph{how} relevant evidence is identified. Structured retrieval via RAG provides a more reliable signal for downstream verification than direct LLM-based extraction, which can blur the boundary between relevance and evidential support.

\begin{table}[t]
\centering
\small
\caption{Ablation of \textsc{DeepSciVerify} pipeline components on the \textsc{SCitance} test set
  (3-class, With NEI). All configurations use GPT-5.4 for abstract-level verification; the extraction method varies in Phase~2, with GPT-4 as the passage-level verifier. $\Delta$ shows Micro-F1 change relative to the abstract-only baseline.
  Best results \textbf{bolded}.}
\label{tab:ablation}
\begin{tabularx}{\columnwidth}{@{} X ccc r @{}}
\toprule
\textbf{Configuration}
  & \textbf{Mi-F1} & \textbf{Ma-F1} & \textbf{S/NS}
  & \textbf{$\Delta$} \\
\midrule
GPT-4 abstract only
  & 82.2 & 78.8 & 86.7 & ref. \\
\addlinespace[2pt]
+ GPT-5.4 LLM extractor
  & 83.3 & 80.8 & 83.3
  & \textcolor{green!50!black}{+1.1} \\
\addlinespace[2pt]
\rowcolor{green!12}+ GPT-5.4 RAG extractor
  & \textbf{86.7} & \textbf{81.5} & \textbf{88.9}
  & \textcolor{green!50!black}{\textbf{+4.5}} \\
\bottomrule
\end{tabularx}
\end{table}

\section{Discussion}

Our results highlight a limitation of single-step verification with large language models. When asked to decide from abstract-level evidence alone, models must both interpret the claim and assess whether the available evidence is sufficient. As shown in Section~\ref{sec:calibration}, this leads to consistent but different behaviors across models: some tend to be conservative and default to \texttt{NEI}, while others are more decisive and risk overcommitting under incomplete evidence. These differences suggest that model outputs are influenced not only by reasoning ability, but also by implicit calibration tendencies.

We also find that performance depends on how evidence is provided. Abstracts are widely available but often lack the detail needed to verify citation-level claims, while naively providing full-text documents leads to only modest improvements (Section~\ref{sec:main_results}). This suggests that the challenge is not only retrieving more information, but selecting the parts of the document most informative for the claim.

Our approach combines these observations. By using a staged design, the system defers uncertain cases at the abstract level and then performs focused verification with passage-level evidence. At the same time, we leverage differences in model behavior by assigning models to roles that better match their tendencies. While our experiments are limited to \textsc{SCitance}, the results suggest that combining complementary model behaviors with selective evidence escalation can improve reliability in scientific verification tasks.

\section{Conclusion}


We introduced \textsc{DeepSciVerify}, a two-stage pipeline for scientific claim--citation verification that uses abstract-level evidence first and selectively escalates to passage-level evidence when the abstract is insufficient. Our analysis shows that LLMs exhibit different calibration behaviors under uncertainty, and \textsc{DeepSciVerify} leverages these complementary tendencies through a staged design. On \textsc{SCitance}, the system outperforms strong abstract-only baselines by +4.5 Micro-F1 in the three-class setting. While our evaluation is limited by the small test set and single-domain benchmark, the results suggest that selective evidence escalation is a promising direction for improving scientific claim--citation alignment when relevant evidence is absent from abstracts but available in the full text.

\bibliography{ref-align}
\bibliographystyle{icml2024}

\newpage
\appendix

\section{Prompts Configuration}
\label{app:prompts}

All LLM-based components are prompted to return structured JSON outputs, enabling deterministic parsing and integration into the retrieval and verification pipeline. Unless otherwise stated, all model calls use temperature $0$. This appendix reports the prompts used for the main comparison modules and the LLM-assisted retrieval fallbacks.

\subsection{Abstract-Level Verification Prompt}
\label{app:abstract_prompt}

The abstract-level verifier is used in Phase~1. Given a claim and the cited paper's abstract, the model predicts whether the abstract \texttt{SUPPORTS}, \texttt{CONTRADICTS}, or provides \texttt{NOT\_ENOUGH\_INFO} for the claim.

\paragraph{System prompt.}
\begin{promptbox}
With a specific abstract, please make an estimation whether the abstract SUPPORTS or CONTRADICTS the claim, or if there is NOT_ENOUGH_INFO to determine.
You must choose SUPPORTS or CONTRADICTS or NOT_ENOUGH_INFO.
Please return your answer as a valid JSON object with the verdict as one of the capitalized tokens (SUPPORTS, CONTRADICTS, NOT_ENOUGH_INFO), as well as an explanation or rationale for the answer.
Respond with ONLY a valid JSON object — no markdown fences, no extra text:
{
  "verdict": "<SUPPORTS | CONTRADICTS | NOT_ENOUGH_INFO>",
  "reasoning": "<explanation or rationale for the answer>"
}
\end{promptbox}

\paragraph{User prompt.}
\begin{promptbox}
Abstract: {abstract}
Claim: {claim}
\end{promptbox}

\subsection{Passage-Level Verification Prompt}
\label{app:passage_prompt}

The passage-level verifier is used in Phase~2 after the pipeline escalates beyond the abstract. The passage is treated as the primary evidence, while the abstract provides supplementary context when available.

\paragraph{System prompt.}
\begin{promptbox}
With a specific passage from a paper and the paper's abstract, please make an estimation whether the evidence SUPPORTS or CONTRADICTS the claim, or if there is NOT_ENOUGH_INFO to determine.
The PASSAGE is the primary evidence. The ABSTRACT provides supplementary context but should not override what the passage says.
You must choose SUPPORTS or CONTRADICTS or NOT_ENOUGH_INFO.
Please return your answer as a valid JSON object with the verdict as one of the capitalized tokens (SUPPORTS, CONTRADICTS, NOT_ENOUGH_INFO), as well as an explanation or rationale for the answer.
Respond with ONLY a valid JSON object — no markdown fences, no extra text:
{
  "verdict": "<SUPPORTS | CONTRADICTS | NOT_ENOUGH_INFO>",
  "reasoning": "<explanation or rationale for the answer>"
}
\end{promptbox}

\paragraph{User prompt.}
\begin{promptbox}
Abstract: {abstract}
Passage: {passage}
Claim: {claim}
\end{promptbox}

When no abstract is available, we use the same verdict schema but omit the abstract field and the instruction that describes the abstract as supplementary context.

\subsection{LLM-Assisted Abstract Search Prompt}
\label{app:abstract_search_prompt}

The abstract retrieval cascade uses LLM-assisted web search only as a final fallback when structured retrieval sources fail. The prompt emphasizes that the model must return only abstracts found from real web results and should prefer a negative result over fabricating a match.

\paragraph{System prompt.}
\begin{promptbox}
You are an academic citation lookup assistant with web search access.

You will receive a CITATION from a scientific research report. Your task is to search the web, find the cited paper, and return its abstract.

Instructions:
1. Use web search to look for the citation — try the title, authors, year, and any URL or identifier provided.
2. Look broadly: the match could be a journal paper, arXiv preprint, conference paper, thesis, technical report, or any other published work.
3. The citation's title may be paraphrased, corrupted, or slightly different from the real title — try searching for key phrases and author names, not just the exact title.
4. When you find the paper, extract its full abstract text.

CRITICAL RULES — read carefully:
- ONLY return information from papers you actually found on the web with a real, working URL that you visited or that appeared in your search results.
- NEVER copy the citation's own text and present it as a "found" result. Every field must come from an independent source you discovered through search.
- If your web search returns NO pages containing this publication, you MUST return {"found": false, ...}. A negative result is correct and expected when a citation cannot be located.
- Do NOT guess, infer, or reconstruct a plausible-sounding abstract. If you are not confident the result is the correct paper, do not include it.
- It is MUCH better to return a negative result than to fabricate a match.

Respond with ONLY a valid JSON object — no markdown fences, no extra text:
{
  "found": true,
  "title": "<exact title as found on the web>",
  "abstract": "<full abstract text>",
  "authors": "<comma-separated author names>",
  "year": "<publication year>",
  "source_url": "<URL where you found the paper>"
}

If you cannot find the paper, respond with:
{
  "found": false,
  "title": "",
  "abstract": "",
  "authors": "",
  "year": "",
  "source_url": ""
}
\end{promptbox}

\paragraph{User prompt.}
\begin{promptbox}
Find the scientific paper described by the citation below and return its abstract.

Citation: {citation}
\end{promptbox}

\subsection{LLM-Assisted Full-Text Search Prompt}
\label{app:fulltext_search_prompt}

The full-text retrieval cascade uses LLM-assisted web search as a final fallback when direct URL, arXiv, Semantic Scholar open-access PDF, and PubMed Central retrieval fail. The model is asked to return a direct, freely accessible full-text URL rather than a landing page.

\paragraph{System prompt.}
\begin{promptbox}
You are an academic paper URL finder with web search access.

You will receive the TITLE of a scientific paper. Your task is to search the web and find a direct link to the paper's full text — a URL that points to a PDF, XML, or HTML version of the complete paper (not just the abstract page).

Preferred sources (in order):
1. PubMed Central (PMC) full text HTML — prefer the HTML article page (e.g. /articles/PMCxxxx/) over the PDF version
2. arXiv PDF or HTML
3. Preprint servers (bioRxiv, medRxiv, SSRN, etc.)
4. Institutional or university repositories
5. Europe PMC (europepmc.org)
6. Open-access mirrors and aggregators
7. Any other freely accessible full-text link

CRITICAL RULES:
- The URL must point to the actual content (PDF, XML, or HTML of the paper), NOT to a landing/abstract page.
- The link MUST be freely and directly accessible without login, CAPTCHA, JavaScript rendering, or any form of authentication. The content will be downloaded programmatically, so it must be accessible via a simple HTTP GET.
- Avoid paywalled publisher websites — they typically block automated downloads even if the page appears accessible in a browser. Prefer open repositories and archives over publisher sites.
- ONLY return URLs you actually found via web search. Do NOT guess or invent URLs.
- If you cannot find a freely accessible direct link, return {"found": false}. Do NOT return a link that requires any form of access control.

Respond with ONLY a valid JSON object — no markdown fences, no extra text:
{
  "found": true,
  "url": "<direct URL to full text PDF/XML/HTML>",
  "format": "<pdf | xml | html>"
}

If you cannot find a link:
{
  "found": false,
  "url": "",
  "format": ""
}
\end{promptbox}

\paragraph{User prompt.}
\begin{promptbox}
Paper title: {title}
\end{promptbox}

\section{Retrieval Cascade Implementation Details}
\label{app:retrieval_details}

This appendix provides additional implementation details for the retrieval modules described in Sections~\ref{sec:abstract_retrieval} and \ref{sec:fulltext_retrieval}. Both retrieval stages use a sequential cascade: stages are attempted in order, and the cascade halts at the first stage whose output satisfies the corresponding acceptance criterion.

\subsection{Citation Parsing}
\label{app:citation_parsing}

Before retrieval, the raw citation string $r$ is parsed into structured retrieval fields. In our implementation, the parser extracts the following fields:
\[
\{\texttt{arxiv\_id}, \texttt{doi}, \texttt{url}, \texttt{title}\}.
\]
These fields determine which retrieval stages can be activated. For example, arXiv-based lookup is attempted only when an arXiv identifier is available, DOI-based retrieval requires a DOI, URL retrieval requires a non-DOI URL, and title-based search requires a parsed title. This parsing step standardizes heterogeneous citation formats into a common representation used by both the abstract and full-text retrieval cascades.

\subsection{Title Similarity Gate}
\label{app:title_similarity}

For abstract retrieval, candidates returned by scholarly APIs are validated using a lightweight title-similarity gate. Given a cited title $t_{\mathrm{cite}}$ and a retrieved title $\hat{t}$, we compute word-overlap similarity:
\begin{equation}
\sigma(\hat{t}, t_{\mathrm{cite}}) =
\frac{|W(\hat{t}) \cap W(t_{\mathrm{cite}})|}
{\max(|W(t_{\mathrm{cite}})|, 1)},
\end{equation}
where $W(t)$ denotes the set of lowercased alphanumeric tokens after removing common English stop words. Unless otherwise stated, we use an acceptance threshold of $\tau = 0.30$. A retrieved abstract is accepted only if it is non-empty and its title similarity satisfies $\sigma \geq \tau$.

\subsection{Abstract Retrieval Cascade}
\label{app:abstract_cascade_details}

The abstract retrieval cascade is organized into four tiers, moving from structured identifiers to broader fallback methods.

\paragraph{Identifier-based retrieval.}
If an arXiv identifier is available, the system first queries the arXiv API and then attempts Semantic Scholar lookup by arXiv ID.

\paragraph{DOI-based retrieval.}
If a DOI is available, the system queries Semantic Scholar and OpenAlex using DOI-based lookup. This stage provides high-precision retrieval when publisher identifiers are present.

\paragraph{Title-based retrieval.}
If identifier-based retrieval fails or no identifiers are available, the system performs title-based search across Semantic Scholar, CrossRef, OpenAlex, and PubMed. Candidate results from these stages are filtered using the title-similarity gate described above.

\paragraph{Fallback retrieval.}
If structured retrieval fails, the system attempts direct URL-based retrieval when a URL is present, followed by LLM-assisted web search as a final fallback. The LLM search prompt requires the model to return only abstracts from papers actually found through web search and to return a negative result rather than fabricate a match.

\subsection{Full-Text Retrieval Cascade}
\label{app:fulltext_cascade_details}

The full-text retrieval cascade is similarly organized into progressively broader stages.

\paragraph{Direct retrieval.}
If the citation contains a URL, the system first attempts to fetch the document directly. HTML pages are converted to text, and PDFs are parsed into text when possible.

\paragraph{arXiv retrieval.}
If an arXiv identifier is available, the system retrieves the arXiv HTML version when available, preferring it because it is typically cleaner for text extraction. If HTML retrieval fails, the system falls back to the arXiv PDF.

\paragraph{Open-access retrieval.}
If earlier stages fail, the system searches open-access sources, including Semantic Scholar open-access PDF links and PubMed Central XML. Retrieved documents are converted into raw text for downstream passage extraction.

\paragraph{Fallback retrieval.}
As a final stage, the system uses LLM-assisted web search to locate a direct full-text URL. The prompt asks for a freely accessible PDF, XML, or HTML version of the paper, prioritizing open repositories such as PubMed Central, arXiv, preprint servers, and institutional repositories. Returned URLs are fetched programmatically and must pass the same quality checks as other full-text sources.

\subsection{Full-Text Acceptance Criteria}
\label{app:fulltext_acceptance}

Full-text candidates are accepted only if they provide sufficient usable content. Specifically, a retrieved document must contain at least $1{,}500$ characters of extracted text and must not match common correction or erratum patterns. We filter documents whose extracted text begins with indicators such as \textit{correction for}, \textit{erratum}, \textit{corrigendum}, \textit{retraction notice}, \textit{author correction}, or \textit{publisher's note}. These checks reduce cases where the retrieved document is not the target paper but rather a correction, notice, or metadata page.

\subsection{Diagnostics}
\label{app:retrieval_diagnostics}

Each retrieval call records the source that produced the final result, the fetched URL when applicable, and a stage-level log of attempted retrieval steps. These logs are used for post-hoc analysis of retrieval failures and for attributing downstream errors to retrieval, passage extraction, or verification.

\section{Model and Experiment Configuration}
\label{app:model_config}

Table~\ref{tab:repro} summarizes the model, retrieval, and reproducibility settings used in the main \textsc{DeepSciVerify} experiments. The main pipeline uses OpenAI models for verification, retrieval fallbacks, and embeddings, while Claude Sonnet 4.6 and Gemini 2.5 Flash are included as baseline models through their respective API providers. All thresholds and retrieval parameters were fixed before evaluation on the \textsc{SCitance} test split.

\begin{table*}[t]
\centering
\caption{Model, retrieval, and reproducibility details for the main \textsc{DeepSciVerify} experiments.}
\label{tab:repro}
\renewcommand{\arraystretch}{1.2}
\small
\begin{tabular}{p{0.24\textwidth}p{0.70\textwidth}}
\toprule
\textbf{Item} & \textbf{Configuration} \\
\midrule

Models used &
The main \textsc{DeepSciVerify} pipeline uses \texttt{gpt-5.4} as the abstract-level verifier in Phase~1 and \texttt{gpt-4} as the passage-level verifier in Phase~2. LLM-assisted retrieval fallbacks use \texttt{gpt-5.2}, and the RAG passage retriever uses \texttt{text-embedding-3-small}. Baseline comparisons additionally use GPT-4, GPT-5.4, Claude Sonnet 4.6, and Gemini 2.5 Flash. \\

\addlinespace[2pt]
API providers &
OpenAI is used for the main \textsc{DeepSciVerify} pipeline, including verifier models, embedding calls, and web-search-augmented retrieval. Anthropic is used for Claude Sonnet 4.6 baselines, and the Google AI API is used for Gemini 2.5 Flash baselines. Non-LLM retrieval uses arXiv, Semantic Scholar, CrossRef, OpenAlex, and NCBI E-utilities for PubMed and PubMed Central. \\

\addlinespace[2pt]
Decoding settings &
All LLM calls use temperature $0$. Reasoning effort is set to \texttt{low} for the abstract-level verifier. Web-search-augmented retrieval calls use the OpenAI \texttt{web\_search\_preview} tool. \\

\addlinespace[2pt]
Abstract retrieval &
The Phase~1 title-similarity gate uses word overlap with threshold $\tau=0.30$. Abstract candidates are accepted only when non-empty and sufficiently title-matched to the parsed citation. \\

\addlinespace[2pt]
Full-text retrieval &
The Phase~2 full-text gate requires at least $L_{\min}=1{,}500$ extracted characters and filters correction, erratum, corrigendum, retraction, author-correction, and publisher-note documents. PDF extraction is capped at $5\times10^{5}$ characters. \\

\addlinespace[2pt]
Passage retrieval &
The RAG passage extractor uses section-aware chunking, chunk size $3{,}000$ characters, overlap $200$, cosine-similarity threshold $0.50$, and top-$k=2$ retrieved chunks. Retrieved chunks are concatenated and passed to the passage-level verifier. \\

\addlinespace[2pt]
Retry and caching &
Per-source rate limiting uses a minimum interval of $\delta=1.0$\,s. Semantic Scholar requests retry up to three times on HTTP 429 with exponential backoff. LLM-assisted abstract and full-text web-search fallbacks allow up to two retries. The pipeline uses an in-memory cache to deduplicate requests within a run, but predictions and retrieval results are not persisted across runs. \\

\addlinespace[2pt]
Experiment period &
Experiments were run between March 23 and March 27, 2026. \\

\bottomrule
\end{tabular}
\end{table*}

\section{Escalation Analysis}
\label{sec:escalation}

Table~\ref{tab:escalation} analyzes how often \textsc{DeepSciVerify} uses each stage. Of the 91 test instances, 61 (67.0\%) are resolved at Phase~1 through early exit, and 57 of these are correct. The remaining 30 instances (33.0\%) are escalated to Phase~2 for passage-level verification.

Escalation has a positive net effect. Phase~2 corrects 13 instances that were initially unresolved at the abstract level and correctly leaves 9 additional cases as \texttt{NEI}. However, it also flips 7 instances to an incorrect verdict, showing that passage-level evidence can introduce noise when retrieval or evidence interpretation is imperfect. Overall, Phase~2 adds 13 correct predictions while introducing 7 new errors, yielding a net gain of 6 correct predictions. This accounts for the +4.5 Micro-F1 improvement over the strongest abstract-only baseline.

\begin{table}[H]
  \centering\small
  \caption{Escalation behavior of \textsc{DeepSciVerify} on the \textsc{SCitance} test set in the three-class setting. Phase~1 performs abstract-level verification, while Phase~2 applies RAG-based passage extraction followed by passage-level verification.}
  \label{tab:escalation}
  \begin{tabularx}{\columnwidth}{@{} X r @{}}
    \toprule
    \textbf{Statistic} & \textbf{Count (\%)} \\
    \midrule
    Total test instances                & 91 (100\%) \\
    \addlinespace[4pt]
    Resolved at Phase~1 (early exit)    & 61 (67.0\%) \\
    \quad Correct                       & 57 (62.6\%) \\
    \quad Incorrect                     & 4 (4.4\%) \\
    \addlinespace[4pt]
    Escalated to Phase~2                & 30 (33.0\%) \\
    \quad Corrected by Phase~2          & 13 (14.3\%) \\
    \quad Remained NEI (correctly)      & 9 (9.9\%) \\
    \quad Remained NEI (incorrectly)    & 1 (1.1\%) \\
    \quad Flipped to wrong verdict      & 7 (7.7\%) \\
    \bottomrule
  \end{tabularx}
\end{table}

\section{Retrieval Coverage and Latency Analysis}
\label{app:retrieval_coverage_latency}

To evaluate the coverage and runtime behavior of the retrieval modules, we ran both the abstract and full-text retrieval cascades on the unique cited papers appearing across the \textsc{SCitance} train, development, and test splits. This analysis uses 412 unique papers after deduplication. Because some papers appear in multiple splits, the per-split totals sum to 527 paper--split occurrences rather than 412 unique papers.

\subsection{Retrieval Latency}
\label{app:retrieval_latency}

Table~\ref{tab:retrieval_latency_app} summarizes retrieval latency. Abstract retrieval is relatively stable, with a median latency of 4.10 seconds per paper and a p95 of 6.89 seconds. Full-text retrieval has a lower median latency of 1.96 seconds, but a substantially heavier tail, with a mean of 12.49 seconds and p95 of 56.70 seconds. This reflects the variability of full-text access: many papers are retrieved quickly from structured sources such as PubMed Central, while fallback web search and document parsing can be much slower.

\begin{table*}[htb]
\centering
\small
\caption{Latency of the retrieval cascades over 412 unique papers.}
\label{tab:retrieval_latency_app}

\begin{tabularx}{\textwidth}{>{\raggedright\arraybackslash}Xccccc}
\toprule
\textbf{Stage} & \textbf{Total time} & \textbf{Median} & \textbf{Mean} & \textbf{P95} & \textbf{Max} \\
\midrule
Abstract retrieval  & 28m37.49s  & 4.10s & 4.17s  & 6.89s  & 9.13s \\
Full-text retrieval & 1h25m46.02s & 1.96s & 12.49s & 56.70s & 97.76s \\
\midrule
End-to-end run      & 1h54m23.53s & -- & -- & -- & -- \\
\bottomrule
\end{tabularx}

\end{table*}

These latency results further motivate the staged design of \textsc{DeepSciVerify}. Since full-text retrieval can be substantially more expensive and variable than abstract retrieval, escalating only uncertain cases reduces unnecessary full-text processing while still enabling deeper verification when abstract evidence is insufficient.

\subsection{Retrieval Coverage}
\label{app:retrieval_coverage}

Table~\ref{tab:retrieval_coverage_app} summarizes overall retrieval coverage. The abstract retrieval cascade successfully retrieves abstracts for 391 of 412 unique papers (94.9\%). Full-text retrieval is more difficult but still succeeds for 333 papers (80.8\%). Only 5 papers (1.2\%) have neither abstract nor full-text evidence retrieved.

\begin{table}[H]
\centering
\small
\caption{Retrieval coverage over 412 unique papers in \textsc{SCitance}.}
\label{tab:retrieval_coverage_app}

\begin{tabularx}{\columnwidth}{>{\raggedright\arraybackslash}Xcc}
\toprule
\textbf{Retrieval outcome} & \textbf{Count} & \textbf{Percentage} \\
\midrule
Abstract retrieved & 391 / 412 & 94.9\% \\
Abstract missing & 21 / 412 & 5.1\% \\
\midrule
Full text retrieved & 333 / 412 & 80.8\% \\
Abstract only & 74 / 412 & 18.0\% \\
No evidence retrieved & 5 / 412 & 1.2\% \\
\bottomrule
\end{tabularx}

\end{table}

Table~\ref{tab:retrieval_sources_app} reports the source distribution for successful retrievals. Abstract retrieval is primarily resolved through PubMed and OpenAlex, while full-text retrieval is dominated by PubMed Central, with LLM-assisted web search recovering an additional 24.0\% of successful full-text cases. This supports the use of a multi-source cascade: no single source covers all cited papers, and the fallback stages recover a non-trivial fraction of evidence.

\begin{table}[H]
\centering
\small
\caption{Source distribution among successful abstract and full-text retrievals. Percentages are computed within each successful retrieval group.}
\label{tab:retrieval_sources_app}

\begin{tabularx}{\columnwidth}{>{\raggedright\arraybackslash}X
                                  >{\raggedright\arraybackslash}X
                                  cc}
\toprule
\textbf{Retrieval type} & \textbf{Source} & \textbf{Count} & \textbf{Share} \\
\midrule
Abstract & PubMed & 192 & 49.1\% \\
Abstract & OpenAlex & 155 & 39.6\% \\
Abstract & CrossRef & 44 & 11.3\% \\
\midrule
Full text & PubMed Central & 253 & 76.0\% \\
Full text & LLM web search & 80 & 24.0\% \\
\bottomrule
\end{tabularx}

\end{table}

For completeness, Table~\ref{tab:retrieval_split_app} reports coverage by split before deduplication across splits. Coverage is broadly consistent across train, development, and test partitions.

\begin{table}[H]
\centering
\small
\caption{Retrieval coverage by \textsc{SCitance} split. Counts are computed over paper--split occurrences, so totals may exceed the number of unique papers.}
\label{tab:retrieval_split_app}
\begin{tabular}{lccccc}
\toprule
\textbf{Split} & \textbf{Papers} & \textbf{Abs. found} & \textbf{Abs. \%} & \textbf{FT found} & \textbf{FT \%} \\
\midrule
Dev & 94 & 88 & 93.6\% & 79 & 84.0\% \\
Test & 85 & 81 & 95.3\% & 73 & 85.9\% \\
Train & 348 & 330 & 94.8\% & 279 & 80.2\% \\
\midrule
Total & 527 & 499 & 94.7\% & 431 & 81.8\% \\
\bottomrule
\end{tabular}
\end{table}

\end{document}